\documentclass[10pt,twocolumn,letterpaper]{article}

\usepackage[pagenumbers]{wacv} 

\usepackage{graphicx}
\usepackage{amsmath}
\usepackage{amssymb}
\usepackage{booktabs}

\usepackage[pagebackref,breaklinks,colorlinks]{hyperref}

\usepackage[capitalize]{cleveref}
\crefname{section}{Sec.}{Secs.}
\Crefname{section}{Section}{Sections}
\Crefname{table}{Table}{Tables}
\crefname{table}{Tab.}{Tabs.}

\def\wacvPaperID{202} 
\def\confName{WACV}
\def\confYear{2025}

\begin{document}

\title{ReinDiffuse: Crafting Physically Plausible Motions with Reinforced Diffusion Model}

\author{Gaoge Han\\ Tencent AI Lab\\
\and
Mingjiang Liang\\
University of Technology Sydney\\
\and
Jinglei Tang\\
Northwest A\&F University\\
\and
Yongkang Cheng\\
Tencent AI Lab\\
\and
Wei Liu\\
University of Technology Sydney\\
\and
Shaoli Huang\thanks{corresponding author}\\
Tencent AI Lab\\
}

\maketitle

\begin{figure*}[t]
    \captionsetup{type=figure}

    \includegraphics[width=1.\textwidth]{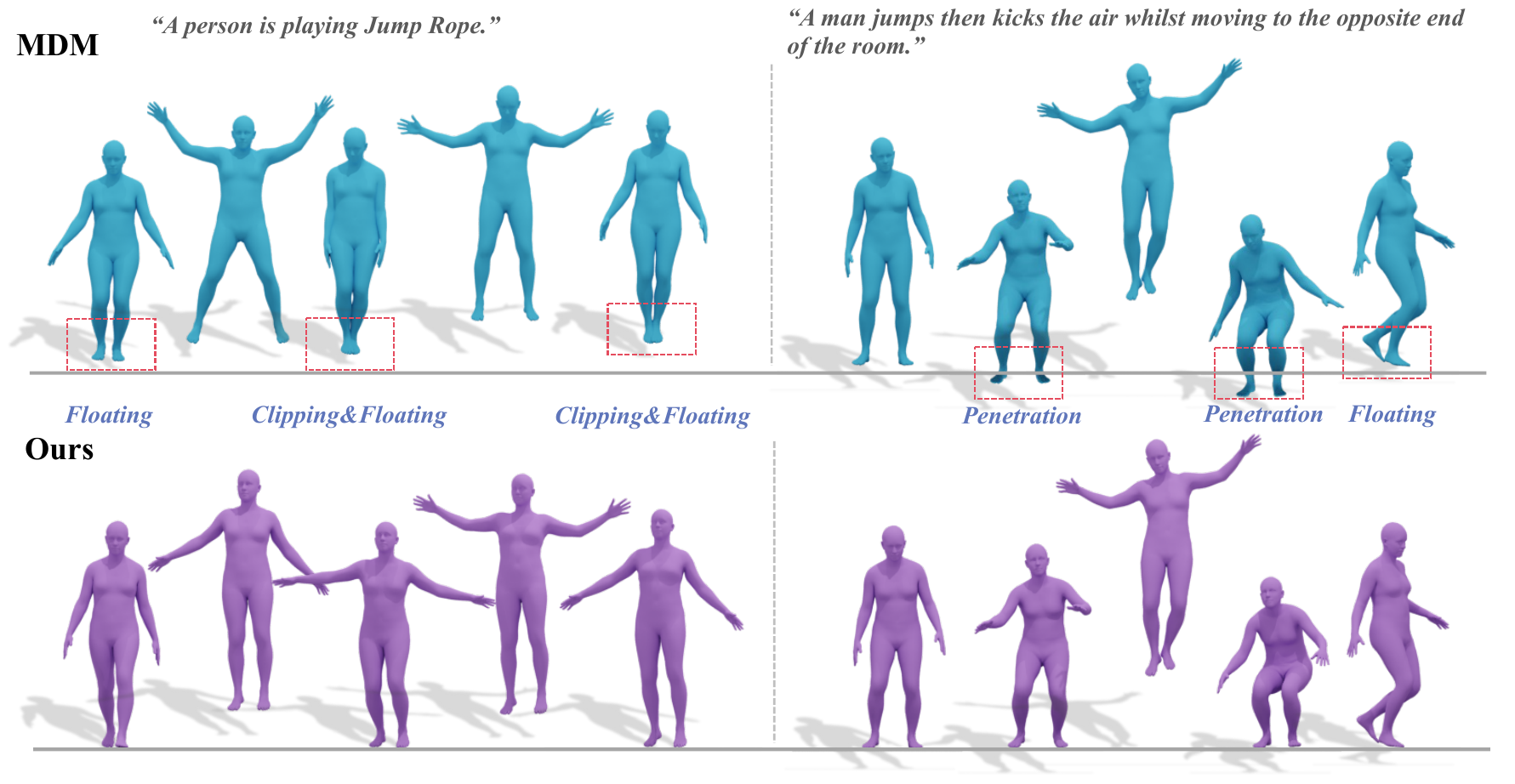}
    \captionof{figure}{ Our ReinDiffuse can generate physically plausible motion, effectively eliminating common physical issues such as floating, penetration, foot clipping, and skating. ReinDiffuse enables MDM to learn physical commonsense with reinforcement learning.  }
    \label{fig:teaser}
\end{figure*}

\begin{abstract}
    Generating human motion from textual descriptions is a challenging task. Existing methods either struggle with physical credibility or are limited by the complexities of physics simulations. In this paper, we present \emph{ReinDiffuse} that combines reinforcement learning with motion diffusion model to generate physically credible human motions that align with textual descriptions. Our method adapts Motion Diffusion Model to output a parameterized distribution of actions, making them compatible with reinforcement learning paradigms. We employ reinforcement learning with the objective of maximizing physically plausible rewards to optimize motion generation for physical fidelity. Our approach outperforms existing state-of-the-art models on two major datasets, HumanML3D and KIT-ML, achieving significant improvements in physical plausibility and motion quality. Project: \url{https://reindiffuse.github.io/}
\end{abstract}

\section{Introduction}
\label{sec:intro}

The quest to generate human motion from textual descriptions has been a frontier in several domains like augmented reality/virtual reality (AR/VR), animation, gaming, and robotics. This pursuit, however, has been marred by a key challenge: producing motions that are not only realistic but also adhere to the stringent laws of physics. This paper introduces an innovative approach that integrates physical commonsense on top of existing motion diffusion models without having to resort to physical simulations.

We unveil a novel method that combines the strengths of reinforcement learning with motion diffusion models to create human motions that are not only physically credible but also align seamlessly with textual descriptions. This approach addresses the critical gap left by existing methods, which either struggle with physical credibility~\cite{mdm2022human} or are bogged down by the complexities of physics simulations~\cite{yuan2023physdiff}.

Our idea is inspired by the success of strong alignment capabilities demonstrated by reinforcement learning (RL) in natural language processing (NLP) and computer vision. InstructGPT~\cite{ouyang2022traininginstructgpt} devises a human feedback reward system to enhance model alignment with human intent using RL. \cite{pinto2023tuning} applies RL to fine-tune computer vision models with task rewards of object detection/segmentation, showing a surprising effectiveness of lifting the alignment between model predictions and intended usage. Drawing from this insight, we explore reinforcement learning methods for improving the alignment between generated motion fidelity and adherence to physical laws for Motion Diffusion Model (MDM) ~\cite{mdm2022human}. However, the incompatibility between RL and MDM and designing reasonable reward functions to incorporate physical commonsense are challenging. 

For the incompatibility problem, our solution hinges on a unique adaptation of Motion Diffusion Model ~\cite{mdm2022human}. Inspired by the classic RL algorithms~\cite{williams1992simpleREINFORCE,schulman2017proximal} and reparameterization techniques~\cite{kingma2013autovae}, we retool these models to output a parameterized distribution of actions, making them compatible with reinforcement learning paradigms. This adaptation is key to bridging the gap between the continuous value output space of MDM and the probability distribution output of reinforcement learning models. For the physical plausibility goal, it is challenging to learn physical laws through the design of reward functions without physical simulation. Considering our goal is to generate physically plausible motion, we do not necessarily need the model to truly understand the physical laws. Instead, we aim to encourage the model to minimize physical issues caused by violations of these laws. With this perspective, we design reward function that focuses on perceiving four types of illogical behaviors that do not have physical validity, including sliding steps, floating, ground penetration and foot clipping.

The heart of our approach is reinforcement learning policy optimization combined with meticulously crafted, physically plausible reward functions. We perform RL policy gradient optimization with importance sampling to maximize the expected reward of motion diffusion model's motion output, which involves two distinct models: a pretrained policy for action sampling and a RL policy that updates based on these samples. This optimization strategy ensures a harmonious balance between exploration and exploitation, crucial for learning realistic motion generation.

Our method's efficacy is validated on two major datasets, HumanML3D \cite{Guo_2022_CVPR_humanml3d} and KIT-ML \cite{Plappert2016kit}. The results are compelling. We achieve a remarkable improvement in physical plausibility and motion quality, significantly outperforming existing state-of-the-art models. Notably, our approach exhibits a notable enhancement in motion quality as evidenced by a 29\% improvement in the Frechet Inception Distance (FID) on the HumanML3D benchmark, and a 34\% improvement on the KIT-ML benchmark. As shown in Fig.~\ref{fig:teaser}, our ReinDiffuse effectively mitigate common physical issues compared to the state-of-the-art MDM~\cite{mdm2022human} methods.

Our contributions are manifold and significant:
\begin{itemize} 

\item A novel approach that synergizes motion diffusion models with reinforcement learning to generate physically credible human motions.
\item An innovative adaptation of MDM for compatibility with reinforcement learning, using parameterized action distributions.
\item A reinforcement learning optimization strategy paired with a physically plausible reward, designed to optimize motion generation for physical fidelity.
\end{itemize}


\section{Related Work}
\label{sec:relatedwork}


{\bf Human Motion Generation.} Human motion generation can be broadly classified into two categories: conditioned and unconditioned. Conditioned inputs include text~\cite{Guo_2022_CVPR_t2m,guo2022tm2t, petrovich22temos,TEACH:3DV:2022,chen2022mld,han2023hutumotion}, music~\cite{lee2019dancing,li2022danceformer,li2021ai, tseng2022edge}, audio~\cite{alexanderson2022listen, zhang2023diffmotion}, images~\cite{rempe2021humor,chen2022learning}, objects~\cite{li2023object}, and trajectories~\cite{wang2021synthesizing, karunratanakul2023GMD,xie2023omnicontrol}. Unconditioned models are represented by~\cite{pavlakos2019expressive,yan2019convolutional,zhang2020perpetual,zhao2020bayesian}.

Recently, extensive research has been conducted on text-driven motion generation models, which can be divided into two main paradigms: Vector Quantized Variational Autoencoder (VQ-VAE)~\cite{van2017neuralvqvae}  based and diffusion-based. Both paradigms demonstrate strong capabilities in diversity and conditional matching. Notable VQ-VAE-based models include T2M-GPT~\cite{zhang2023t2mgpt} and MotionGPT~\cite{jiang2023motiongpt}, while diffusion-based models feature MotionDiffuse~\cite{zhang2022motiondiffuse}, MDM~\cite{mdm2022human}, and ReMoDiffuse~\cite{zhang2023remodiffuse}.

T2M-GPT~\cite{zhang2023t2mgpt} proposes a two-stage motion training method that incorporates convolution-based VQ-VAE  and Generative Pretrained Transformer(GPT). MotionGPT~\cite{jiang2023motiongpt} learns a motion-specific VQ-VAE model that converts raw motion data into a sequence of motion tokens and introduces a pretrained language model for motion-relevant generation tasks. MotionDiffuse~\cite{zhang2022motiondiffuse} is the first diffusion-based text-driven motion generation framework, supporting multi-level manipulation to handle fine-grained text descriptions of body parts. MDM~\cite{mdm2022human} predicts instead of noise to apply the geometric loss in the multi-step diffusion process. ReMoDiffuse~\cite{zhang2023remodiffuse} extends existing motion models with retrieval capabilities to enhance generalizability and diversity.

While current motion generation models have made significant progress in conditioned matching and diversity, they still face challenges in incorporating physical commonsense, as they are primarily designed to fit data distributions. This limitation leads to the generation of physically unrealistic motions.

{\noindent \bf Physically Plausible Motion Generation.} Generating physically plausible motion has always been a challenging task. In 3D scenes, most models~\cite{chao2021learning, pan2023synthesizing, hassan2023synthesizing} aim to create realistic human motions by combining motion control, physics-based simulations~\cite{makoviychuk2021isaac}, and reinforcement learning. In text-to-motion tasks, most models employ transformer-based architectures to fit mocap data and enhance the physical plausibility of generated motion using geometric losses such as velocity and foot contact loss. Although these methods can produce motion that visually corresponds to text descriptions, they often struggle to capture the nuances of realistic human movement, resulting in animations that may appear awkward or unnatural. These shortcomings stem from the models' inability to fully incorporate the complexities of real-world physics into the generated motions.

PhysDiff~\cite{yuan2023physdiff} proposes an approach to generate physically plausible motion in text-to-motion tasks by combining diffusion models and physics simulation. Specifically, it iteratively applies a physics-based motion projection module trained in the Isaac-Gym~\cite{makoviychuk2021isaac} physics simulator during the inference stage of the diffusion model. Although these approaches are more successful in generating movements that adhere to the laws of physics, they come with their own set of challenges. Physics-based simulations are often computationally intensive and complex, requiring significant resources to accurately simulate aspects like gravity, momentum, and material interactions. This complexity can make the models slow, difficult to train, and challenging to integrate into real-time applications such as gaming or VR/AR. 

In this paper, we explore the possibility of enabling the model itself to incorporate physical commonsense to generate physically plausible motion, rather than relying on guidance from training a physics-based motion projection model in physical simulation like PhysDiff~\cite{yuan2023physdiff}. Simultaneously, we aim to eliminate dependence on physics simulators, as text-to-motion tasks cannot be trained directly within a physics simulator.

{\noindent \bf Reinforcement Learning in Artificial Intelligence Generated Content (AIGC).} Recently, the significant success ~\cite{ouyang2022traininginstructgpt, wu2023fine} of Reinforcement Learning from Human Feedback (RLHF) in Large Language Model (LLM) fine-tuning has garnered considerable attention from researchers. It has inspired researchers to explore the extension of RL applications to other AIGC-related tasks. Fine-grained RLHF ~\cite{wu2023fine} incorporates multiple reward models associated with different human feedback types, which can effectively address the problems of detoxification and long-form question answering. ImageReward ~\cite{xu2023imagereward} presents the first general-purpose text-to-image human preference reward model and tunes diffusion models regarding human preference scorers. DPOK~\cite{fan2023dpok} proposes online reinforcement learning to fine-tune pretrained text-to-image diffusion models using feedback-trained reward on the denoising diffusion stage. DDPO~\cite{black2023ddpo} adapts text-to-image diffusion models for directly optimizing diffusion model using reinforcement learning with downstream objectives, such as vision-language model feedback, aesthetic quality, compressibility and incompressibility. Inspired by the success of RL in AIGC, we propose a method that enables direct RL training on motion diffusion model. Our goal is to facilitate the model's ability to avoid physical implausibility by maximizing physical Plausibility rewards.

\section{Preliminaries}
\label{sec:preliminaries}
In this section, we provide a brief background on motion diffusion model and the reinforcement learning (RL) problem formulation.
\subsection{Motion Diffusion Model}
Motion diffusion model (MDM)~\cite{mdm2022human} designs a encoder-only transformer network. Unlike other diffusion-based models,  given the noise time-step $t$ and the condition $c$ and a noised motion $\epsilon_t$, MDM outputs the human motion $x_0$ rather than the noise in each diffusion step. This allows MDM to supervise geometric losses (locations, velocities and foot contact) at each step of diffusion. In the absence of omitted geometric losses, the objective function of MDM can be expressed as:
\begin{equation}
\label{eq:simple}
{
\mathcal{L}_\text{simple} = \mathbb{E}_{x_0 \sim D, t \sim [1,T]}[\| x_0 - G(\epsilon_t, t, c)\|_2^2],
}
\end{equation}
where $G$ denotes the encoder-only transformer model. $D$ denotes the motion data distribution.

Although MDM's performance in semantic matching is less than satisfactory, multiple applications of geometric losses render MDM still the state-of-the-art model in terms of physical plausibility. Therefore, to explore the upper limits of the model's physical performance, we use MDM as our baseline and conduct reinforcement learning training based on MDM.

\subsection{ Reinforcement Learning}
The objective of reinforcement learning (RL) is to enable an agent to learn through interaction with its environment in order to achieve the maximum cumulative reward for its policy network $\pi$ in a given task. The objective $\mathcal{L}_\text{RL}$ can be defined as:
\begin{equation}
{
\mathcal{L}_\text{RL}  = \mathbb{E}_{\tau \sim p( \cdot|\pi)} [ R(\tau) ],
}
\end{equation}
where $R$ denotes the reward function, $\tau$ denotes the trajectories when agent acts in the Markov decision process (MDP). In the fields of gaming and robotics, reinforcement learning can be defined as MDP. When RL is applied to the LLM (e.g. RLHF~\cite{ouyang2022traininginstructgpt}) and Motion Diffusion Model, it can be simplified to a single step (called as the multi-armed bandit problem~\cite{katehakis1987multi,lu2010contextual}). Reinforcement learning relies on the model's output as likelihood for updates, posing a challenge when directly conducting RL training on the MDM model. A well-designed reward to improve physically plausibility is also crucial for reinforcement learning training in MDM.

\begin{figure*}[t]
\centering
\includegraphics[width=\textwidth]{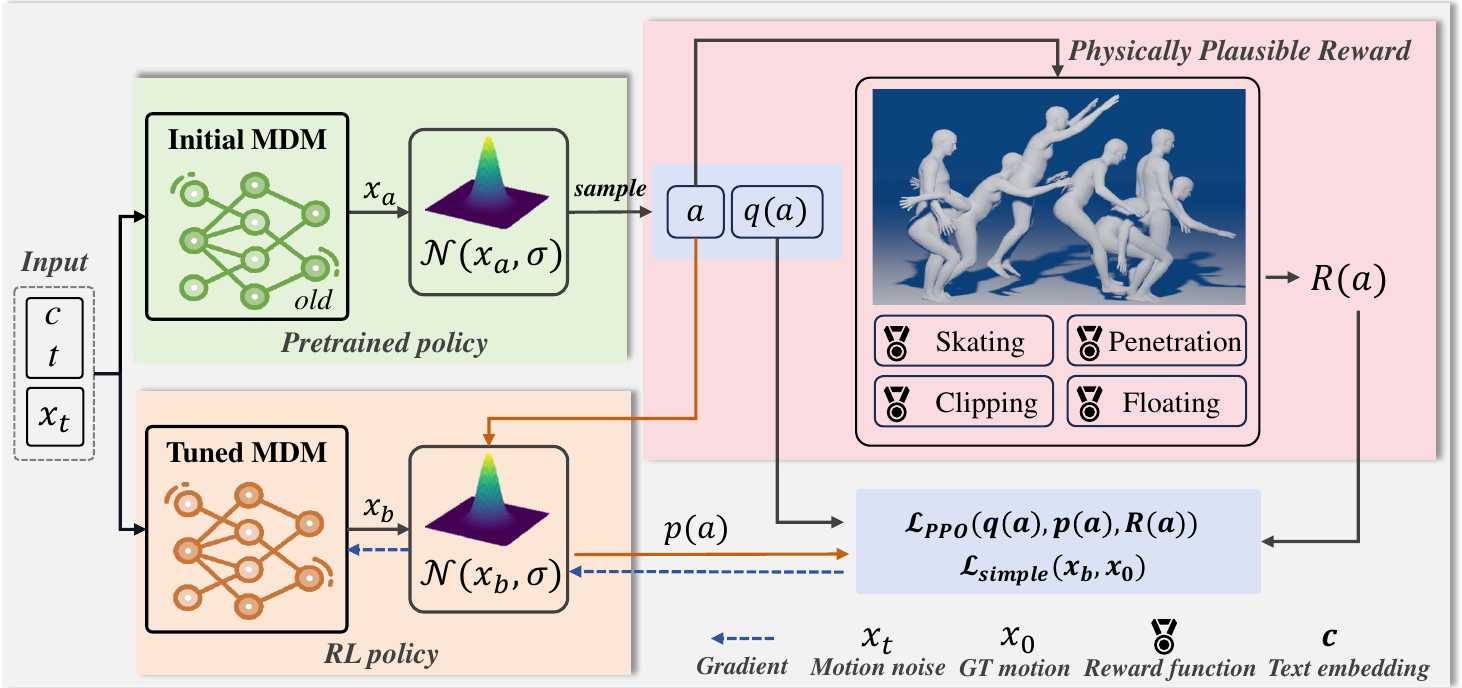}
    \captionof{figure}{ 
Overview of our ReinDiffuse training framework. Given the condition $c$, time step $t$ and noised motion $\epsilon_t$. Initial MDM output the $x_a$ as the mean of Gaussian distribution and use the fixed $\sigma$ as the standard deviation. We sample $a$ from the distribution to compute the physically plausible rewards and obtain the motion likelihood $q(a)$ and $p(a)$. We perform RL training using the combined loss of $\mathcal{L}_{PPO}$ and $\mathcal{L}_{simple}$ with PPO to update tuned MDM.
}
\label{fig:overview}
\end{figure*}

\section{Method}
\label{sec:method}
Our methodology combines motion diffusion model (MDM) and reinforcement learning (RL) to generate realistic human motion based on textual descriptions. This integration is made possible through reparameterization techniques ~\cite{kingma2013autovae} and the utilization of pretrained MDM for initializing policy models. The primary focus is to adapt the model to effectively harness reinforcement learning and the elaborated reward function design for physical plausibility, which ensures the generation of physically plausible motions.

\subsection{Adapting MDM with Reparameterization}
The characteristic of MDM~\cite{mdm2022human} is that it predicts $x_0$ at each diffusion step rather than noise, facilitating the use of established locations and velocities losses. MDM is still the best T2M model in terms of physical performance in practical use to date. MDM inherently generates data in a continuous value format that is not immediately compatible with standard reinforcement learning paradigms. This incompatibility arises because reinforcement learning typically requires a policy output that represents a probability distribution of actions, whereas MDM do not naturally provide their outputs in such a probabilistic format. 

Inspired by the the construction of policy networks of classic RL algorithms~\cite{williams1992simpleREINFORCE,schulman2017proximal} and reparameterization techniques~\cite{kingma2013autovae}, we reparameterize the output of MDM without requiring any dedicated modifications to the model architecture. Specifically, we treat the output of the MDM as the mean of a Gaussian distribution and introduce a predefined standard deviation. This is represented as:

\begin{equation}
   a = \mu(x_0) + \sigma\cdot \upsilon,  
\end{equation}   
where $\upsilon$ is an independent noise variable, $\mu(x_0)$ represents the mean of the Gaussian distribution, derived from the MDM output $x_0=G_{\theta}(\epsilon_t, t, c)$. The term $\sigma$ is a predefined standard deviation, typically sampled from a standard normal distribution. This approach effectively converts the MDM output into a policy format compatible with reinforcement learning methods. By constructing a Gaussian distribution in this manner, the model's output becomes a set of probable actions, each with a calculable likelihood, allowing for the application of reinforcement learning techniques to optimize the policy.

Furthermore, an important aspect of our approach is the initialization of these policy models with pretrained models. By using models that have already been trained on relevant motion data, we provide a knowledgeable starting point for the policy, enhancing the efficiency and effectiveness of the learning process. This pretrained initialization is particularly beneficial in grounding the stochastic outputs of the MDM within a realistic and physically plausible space.
Finally, the reparameterized output $x$ is then used as the policy in the reinforcement learning paradigm. This policy, denoted as $\pi_{\theta}(a|\epsilon_t,t,c,\sigma)$, represents the probability of taking action $a$ given the observation $(\epsilon_t,t,c,\sigma)$.

\subsection{Reinforcement Learning Training}

In a vanilla reinforcement learning framework, a policy model is responsible for both actions sampling and gradient updating itself based on the reward. This paradigm can lead to instability in training, because it tends to learn high correlation between action selection and policy gradient update, reducing exploration and potentially converging to suboptimal solutions. To mitigate this issue, we employ policy gradient optimization with importance sampling and start from the pretrained model, inspired by ~\cite{kakade2002approximately, ouyang2022traininginstructgpt}. As shown in Fig.~\ref{fig:overview}, the pretrained policy denoted as $\pi_{\theta^{PT}}(a|\epsilon_t,t,c,\sigma)$, is primarily responsible for sampling actions based on adapted MDM distribution. It generates a variety of actions and computes the physically plausible rewards for these actions. This policy acts as a broad-spectrum explorer, providing a diverse set of action-reward pairs. The RL policy and denoted as $\pi_{\theta^{RL}}(a|\epsilon_t,t,c,\sigma)$, is focused on learning and gradient optimization. It uses the actions and reward scores sampled by the pretrained policy to update and refine its own strategy. This is where the principle of importance sampling comes into play. The RL policy weights the updates based on the importance or utility of the sampled actions, leading to more informed and stable policy gradient updates. During training, the RL policy undergoes multiple steps of updates, refining its decision-making process. These updates are based on the reward from the diverse set of actions sampled by the pretrained policy, ensuring a balanced approach between exploration and exploitation.

\subsection{Reward Functions for Physical Plausibility}
It goes without saying that the design of a physically plausible reward function in the context of Motion Diffusion Model (MDM) is crucial for ensuring that the generated human motion sequences are realistic and adhere to the laws of physics. To address common issues where the generated motions might violate physical laws, particularly in the context of bipedal movements, we focus on identifying four key non-physical behaviors.

\noindent\textbf{Sliding Steps.} This occurs when the model generates motions where the feet appear to slide on the surface instead of showing clear lifting and placing, resembling a slippery motion. The reward function penalizes motions where foot movements do not correlate with the expected locomotion, ensuring that steps are physically grounded and exhibit natural gait patterns. Thus, given an action $a$, the $i^{th}$ frame's reward function can be expressed as:
 \begin{align}
 R^i_\texttt{s}(a) = \exp(-||(a_{ft}^i-a_{ft}^{i-1})\cdot f^{i}_S\cdot f^{i-1}_{S}||_{2}),  
 \end{align}
where $a_{ft}^i,(f^i_S, f^{i-1}_S)$ represents the $i^{th}$ foot joint locations and the contact label, respectively. 

\noindent\textbf{Floating.} Floating happens when the feet are animated as if they are not making proper contact with the ground, giving an impression of hovering or floating. The reward system is designed to detect and penalize instances where the feet do not maintain realistic contact with the ground, promoting proper weight bearing and balance in the motion sequences.
\begin{align}
R^i_\texttt{F}(a) = \exp(-||(a_{h}^i - h_{ground}) \cdot f^i_{F}||_{2}),
\end{align}
where $a_{h}^i$ denotes the lowest joint height of $i^{th}$ frame, $h_{ground}$ denotes the height of ground. The default $f^i_{F} = 0$. When $a_{h}^i > h_{ground}$, $f^i_{F} = 1$.

\noindent\textbf{Ground Penetration.} This refers to scenarios where parts of the feet or legs incorrectly intersect with the ground plane, breaking the illusion of physical solidity. The function reduces rewards for any motions where there is unnatural intersection or penetration of the feet with the ground surface, ensuring the integrity of physical interactions. We denote the reward function as:
\begin{align}
R^i_\texttt{P}(a) = \exp(-||(h_{ground} - a_{h}^i ) \cdot f^i_{P}||_{2}),
\end{align}
where the default $f^i_{F} = 0$. When $a_{h}^i < h_{ground}$, $f^i_{F} = 1$.

\noindent\textbf{Foot Clipping.} Foot Clipping occurs when both feet of the character model intersect with each other in an unnatural way. The reward function identifies and penalizes scenarios where the character's feet intersect or penetrate each other, maintaining the physical plausibility of body movements and postures, which is given as:
\begin{align}
R^i_\texttt{C}(a) = \exp(-||(a_{lf}^i - a_{rf}^i) \cdot f^i_{C}||_{2}),
\end{align}
where $a_{lf}^i$ and $a_{rf}^i$ represent the left foot and right foot in $i^{th}$ frame, respectively. When the distance between foot less than a threshold, the $f^i_{C}$ is equal to 1. otherwise, it is 0.

 These reward functions are integrated into the reinforcement learning optimization with the objective of maximizing cumulative physical credibility rewards in a frame-by-frame manner. It's worth noting that the reward function we designed is essentially a piecewise function, which is non-differentiable. Therefore, it cannot be used for fine-tuning in supervised learning. The total reward $R^i(a)$ for the $i^{th}$ frame is calculated by accumulating the four rewards. It jointly evaluates the generated motions based on these physical criteria and provides feedback that guides the optimization of the models. In addition, our reward functions are designed to calculate rewards on a frame-by-frame basis within each generated motion sequence. This provides a more granular and precise assessment of the physical plausibility, making it easier to pinpoint and correct non-physical behaviors quickly.

\subsection{Loss Functions and Optimization}

In our methodology, we optimize the policy using two distinct objective functions: one based on Proximal Policy Optimization algorithm (PPO)~\cite{schulman2017proximal} $\mathcal{L}_{PPO}$ and another inherent to the Motion Diffusion Model (MDM)~\cite{mdm2022human} $\mathcal{L}_{simple}$ that is defined at Eq.~\ref{eq:simple}. The PPO objective function is designed to refine the policy within the reinforcement learning framework:
\vspace{-5pt}
\begin{equation}
  \mathcal{L}_{PPO}(\theta) = \mathbb{E}\left[\min(r_i(\theta)R_i, \text{clip}(r_i(\theta), 1-\gamma, 1+\gamma)R_i)\right], 
\end{equation}
where, $r_i(\theta)=\frac{\pi_{\theta^{RL}}(a|\epsilon_t,t,c,\sigma)}{\pi_{\theta^{PT}}(a|\epsilon_t,t,c,\sigma)}$represents the probability ratio between the RL policy and the pretrain policy as introduced previously, $R_i$ is the reward function at the $i^{th}$ frame, and $\gamma$ is a hyperparameter for clipping. Finally, we train the framework using the combined object loss $\mathcal{L}=\mathcal{L}_{PPO} +  \lambda \mathcal{L}_{simple}$, which ensures a comprehensive optimization that not only enhances the reinforcement learning process but also maintains the capability of MDM.

\setlength{\tabcolsep}{3pt}
\begin{table}[t]

\footnotesize
\centering
\resizebox{0.95\linewidth}{!}{ 
\begin{tabular}{lcccccc}
\toprule
Method       & FID$\downarrow$ & R-Precision$\uparrow$ & Skate ratio$\rightarrow $ & Float (m)$\rightarrow$  & Penetrate (m)$\downarrow$  & Clip (m)$\downarrow$ \\ \midrule

Real               & 0.002         & 0.797          & 0.057            & 0.704             & 0.000             & 0.000        
\\ \midrule
T2M~\cite{guo2022t2m}     & 1.067          & 0.740 & 0.217          & 3.368          & 0.235          & 0.263           \\
MotionDiffuse ~\cite{zhang2022motiondiffuse} & 0.630  & \textbf{0.782} &  0.426          & 2.545          & 0.386          & 0.217       \\
MDM~\cite{mdm2022human}   & 0.544   & 0.611   & 0.102       & 1.757       & 0.048    & 0.014      \\

PhysDiff~\cite{yuan2023physdiff}  & 0.433 & 0.631   & - & - & - & - \\
\midrule
ReinDiffuse (Ours)   & \textbf{0.385} & 0.622 & \textbf{0.058} & \textbf{0.711} & \textbf{0.000} & \textbf{0.000}\\

\bottomrule
\end{tabular}
}
\caption{Text-to-motion results on HumanML3D~\cite{Guo_2022_CVPR_humanml3d}. $\rightarrow$ means closer to real is better. \textbf{Bold} indicate the best results.}
\label{table:humanml3d}

\end{table}

\setlength{\tabcolsep}{3pt}

\begin{table}[t]
\footnotesize
\centering
\resizebox{.95\linewidth}{!}{ 
\begin{tabular}{lcccccc}
\toprule
Method       & FID$\downarrow$ & R-Precision$\uparrow$ & Skate ratio$\rightarrow$ & Float (m)$\rightarrow$  & Penetrate (m)$\downarrow$  & Clip (m)$\downarrow$ \\ \midrule

Real               &  0.031      &  0.779 & 0.085   & 0.930             & 0.000             &0.000          
\\ \midrule
T2M~\cite{guo2022t2m}                      & 2.770          & \textbf{0.693}   & 0.364  & 3.782        & 0.189        & 0.283       \\
MDM~\cite{mdm2022human}                         & 0.497          & 0.396       & 0.109    & 1.509       & 0.023       & 0.094      \\

\midrule
ReinDiffuse (Ours)          & \textbf{0.326}     & 0.405 & \textbf{0.087} & \textbf{0.938} & \textbf{0.000} & \textbf{0.000} \\

\bottomrule
\end{tabular}
}
\caption{Text-to-motion results on and KIT-ML~\cite{Plappert2016kit}. $\rightarrow$ means closer to real is better. \textbf{Bold} indicate the best results.}
\label{table:kit}
\end{table}

\section{Experiments}
\label{sec:experiments}

To evaluate the ability to generate physically plausible motions, we conduct text-to-motion experiments using two large-scale datasets: HumanML3D \cite{Guo_2022_CVPR_humanml3d} and KIT \cite{Plappert2016kit}. In Sections \ref{subsec:datasets} and \ref{subsec:implementation}, we present the datasets, evaluation metrics, and implementation details. In Section \ref{subsec:comparison}, we compare our quantitative and qualitative results with state-of-the-art physically plausible methods (MDM \cite{mdm2022human}, PhysDiff \cite{yuan2023physdiff}) to address the following questions: (1) Can our end-to-end reinforcement learning training enable the model to capture physical commonsense? (2) Can our approach surpass PhysDiff (i.e., training an extra motion projection model in a physical simulation environment) in generating higher-quality motion? In Section \ref{subsec:ablation}, we conduct ablation studies to explore the impact of our method's hyperparameters.

\subsection{Datasets and Evaluation Metric}
\label{subsec:datasets}
We provide a brief introduction to two well-known datasets for text-driven motion generation: HumanML3D \cite{Guo_2022_CVPR_humanml3d} and KIT-ML \cite{Plappert2016kit}. Our evaluation metrics focus on three main aspects: physical plausibility, motion quality, and text-conditioned matching.

{\bf Datasets.} \textbf{HumanML3D} dataset \cite{Guo_2022_CVPR_humanml3d} contains 14,616 human motion sequences and 44,970 text descriptions, comprising 5,371 distinct words. Each motion is paired with at least three textual descriptions. The average motion length is 7.1 seconds, with a minimum duration of 2 seconds and a maximum of 10 seconds. All motions are rescaled to 20 FPS. \textbf{KIT-ML} dataset \cite{Plappert2016kit} contains 3,911 human motion sequences and 6,278 textual annotations, averaging 9.5 words per description. Each motion corresponds to multiple text descriptions, ranging from one to four. The dataset has a total vocabulary size of 1,623 unique words, disregarding capitalization and punctuation. All pose sequences are downsampled to 12.5 FPS.

{\bf Evaluation Metrics.} 
For physical plausibility metrics, we report the foot skating ratio (Skate ratio), ground floating (Float), ground penetration (Penetrate), and foot clipping (Clip). For each of these metrics, we aggregate all physical issue values within the motion and report the averages across all samples, excluding foot skate ratios. Following GMD \cite{karunratanakul2023GMD}, the foot skating ratio quantifies the percentage of frames in which either foot skids more than a specified distance (2.5 cm) while maintaining contact with the ground (foot height $<$ 5 cm). Ground floating measures the distance between the ground and the lowest joint positions above the ground ($>$ 5 cm). Ground penetration measures the distance between the ground and the lowest joint positions below the ground ($<$ 0 cm). Foot clipping measures the distance between the left and right feet when it is less than a certain distance threshold (5 cm). For motion quality, we adopt the Fréchet Inception Distance (FID) \cite{heusel2017gansfid}, which calculates the distribution distance between the generated and real motion under motion features. For text-conditioned matching, we adopt the R Precision Top 3 metric that calculates the text and motion embedding \cite{Guo_2022_CVPR_humanml3d} Top 3 matching accuracy based on Euclidean distances.

\subsection{Implementation Details}
\label{subsec:implementation}
We implement our method based on MDM \cite{mdm2022human} using Pytorch and train it on a single NVIDIA A40 GPU. Proximal policy optimization (PPO)~\cite{schulman2017proximal} is used to train the RL policy. For the hyperparameters of our RL training, we conduct a meticulous grid search on the test set to determine the most optimal values. The $\lambda$ is set to 0.4, the PPO clipping coefficient $\gamma$ is set to 0.2, the $\sigma$ is set to 0.15, and the learning rate is 1e-7, while retaining the default hyperparameters in AdamW. We fine-tune the MDM using our method for 50K steps. At the end of each iteration, we update the policy by iterating over the samples for 4 inner epochs with a batch size of 32. The other settings remain the same as in MDM. In the inference stage, we directly use the model's output $x_0$ to remove the fixed standard deviation and achieve better motion generation performance.

\begin{figure*}[t]
    \captionsetup{type=figure}
    \includegraphics[width=\textwidth]{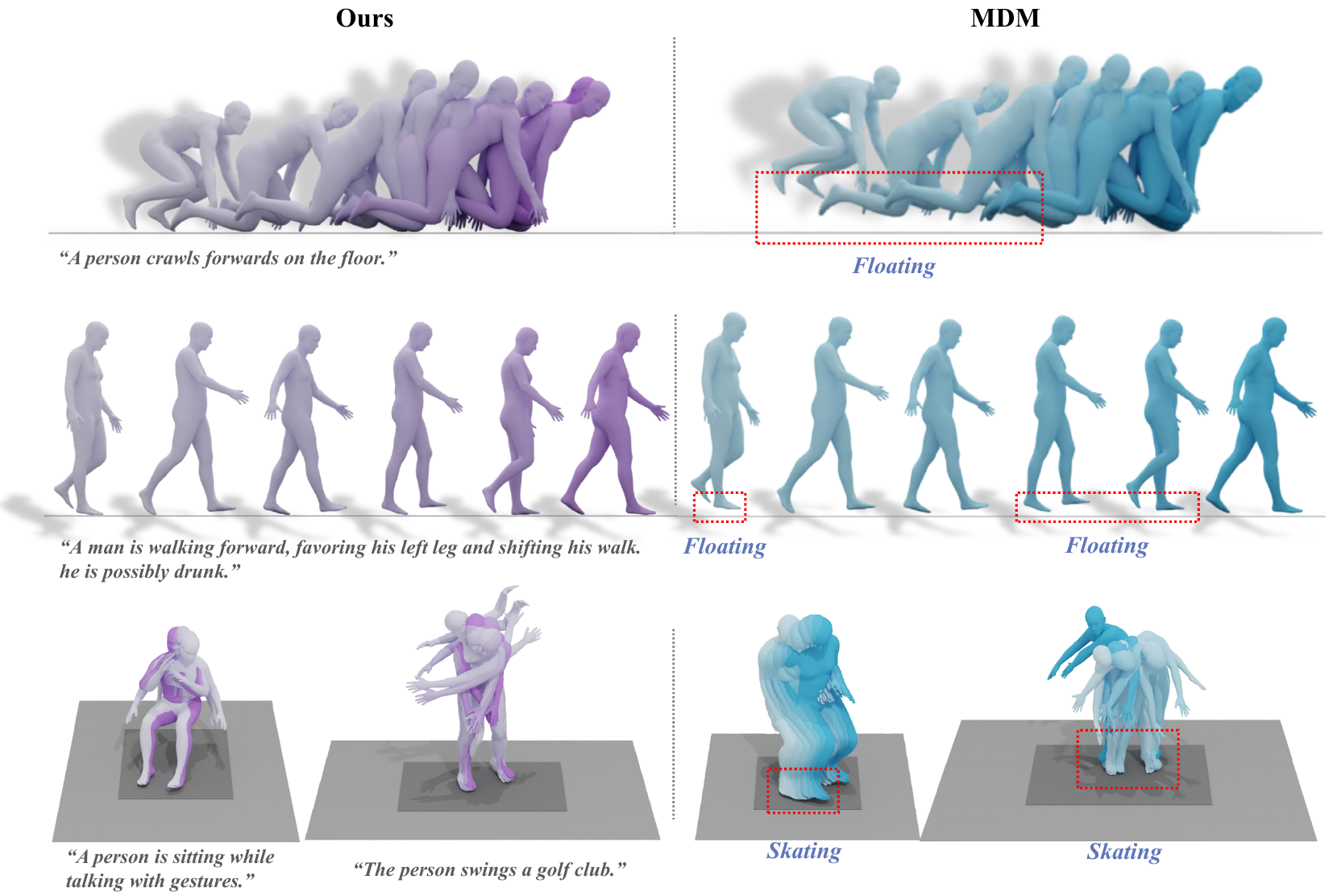}
    \captionof{figure}{ 
 Visual results of ReinDiffuse against the MDM. The darker colors indicate the later frame in time.}
 \label{fig:vis}
\end{figure*}

\subsection{Comparison to State-of-the-art Approaches}
\label{subsec:comparison}
In this section, we present qualitative and quantitative comparison results on the HumanML3D \cite{Guo_2022_CVPR_humanml3d} and KIT-ML \cite{Plappert2016kit} test sets, comparing our method to existing state-of-the-art approaches \cite{mdm2022human, yuan2023physdiff}. It is worth noting that MDM currently has the best physical performance for text-conditioned motion generation models, thanks to the multiple applications of geometric loss during the diffusion process. To explore the upper limit of physically plausible generation, we conducted fine-tuning experiments on physics rewards using MDM as the baseline. Furthermore, PhysDiff is a physically reliable motion generation method based on MDM. Due to the unavailability of PhysDiff's source code, our comparison was limited to the FID metric using our four physical metrics. We reproduce the quantitative comparison results compared our method to PhysDiff using PhysDiff's physical metrics calculation in the  supplementary material.

{\bf Quantitative Comparison.}
In Table \ref{table:humanml3d} and Table \ref{table:kit}, we present the quantitative comparison results for the HumanML3D \cite{Guo_2022_CVPR_humanml3d} and KIT-ML \cite{Plappert2016kit} test sets. On both datasets, our model exhibits physical plausibility close to real performance across Skate ratio, Float, Penetrate, and Clip metrics. This indicates that our RL fine-tuning approach, which aims to maximize physical law rewards, effectively enables MDM to incorporate physical commonsense. From the perspectives of FID and R-Precision metrics, our approach significantly improves the FID metrics by 29\% and 34\% over MDM without compromising the semantic matching capability on HumanML3D and KIT-ML, respectively. In fact, there is even a slight improvement in R-Precision. These findings demonstrate the compatibility of our method, improving motion quality without compromising other model capabilities. Furthermore, it indicates the robustness of our method across datasets of varying sizes.

While a direct comparison of physical metrics with PhysDiff is not feasible, our superior motion quality is reflected in the FID scores. PhysDiff requires applying 50 steps of motion projection during the inference stage of MDM, resulting in a 2-3 times slower motion generation compared to MDM. In contrast, our end-to-end RL training approach, which removes the standard deviation of $\sigma$ during inference and directly utilizes model outputs, ensures that our method does not decrease MDM's inference speed.

\begin{table}[t]

\footnotesize
\centering
\resizebox{0.95\linewidth}{!}{ 
\begin{tabular}{lcccccc}
\toprule
Method       & FID$\downarrow$ & R-Precision$\uparrow$ & Skate ratio$\rightarrow $ & Float (m)$\rightarrow$  & Penetrate (m)$\downarrow$  & Clip (m)$\rightarrow$ \\ \midrule

$\sigma = 0.1$        & 0.392      & 0.621 &0.059  & 0.726 & \textbf{0.000}                 &  \textbf{0.000}   \\
$\sigma = 0.15$& \textbf{0.385} & 0.622 & \textbf{0.058} & \textbf{0.711}  & \textbf{0.000} & \textbf{0.000}\\
$\sigma = 0.2$           &   0.395 & 0.612  & 0.061 & 0.740 & 0.006                          & 0.002    \\
$\sigma = 0.25$   &  0.412 & \textbf{0.624} & 0.065 & 0.782 & 0.016                          & 0.003\\
$\sigma = 0.3$           &   0.454 & 0.616 & 0.071 & 0.835 & 0.021                           & 0.005\\

\bottomrule
\end{tabular}
}
\caption{Effect of varying standard deviation $\sigma$ in RL fine-tuning on HumanML3D~\cite{Guo_2022_CVPR_humanml3d}.}
\label{table:sigma}
\end{table}

{\bf Qualitative Comparison.}
In Fig.~\ref{fig:vis}, we present qualitative results compared to MDM~\cite{mdm2022human} on the HumanML3D~\cite{Guo_2022_CVPR_humanml3d} and KIT-ML~\cite{Plappert2016kit} test sets. Visual comparisons clearly illustrate a substantial improvement in the physical plausibility of MDM following RL fine-tuning with the objective of maximizing physical law rewards. In the first row, the input prompt is ``A person crawls forwards on the floor." Although the generated motion is semantically correct, we observe a significant issue of floating in the initial crawling action. In the second row, the input prompt is ``A man is walking forward, favoring his left leg and shifting his walk. He is possibly drunk." We can see that the motion generated by MDM correctly reflects the textual description ``favoring his left leg," and most of the actions have fewer physical issues. However, there are still noticeable floating problems at the beginning and end. In comparison to MDM's results, our approach generates motion for these two prompts that is both semantically correct and physically plausible, eliminating the issue of floating. In the third row, we present prompts, namely, ``A person is sitting while talking with gestures." and ``The person swings a golf club." along with motions generated by our method and MDM. By observing the distance of foot movement from the initial to the final frame, it is evident that MDM exhibits significant sliding issues. In contrast, the motion generated by our approach does not exhibit foot sliding problems.

\subsection{Ablation Studies}
\label{subsec:ablation}
In this section, we conduct ablation studies to explore the effect of varying standard deviation $\sigma$ and the effect of importance sampling technique (see the supplementary material for details) in RL training for MDM on the HumanML3D~\cite{Guo_2022_CVPR_humanml3d} dataset. 

{\bf The Effect of Varying Standard Deviation $\sigma$.} By representing the output of MDM as the mean of a Gaussian distribution and employing a fixed standard deviation, we can obtain motion likelihood for reinforcement learning training using PPO. The use of a fixed variance aims to simplify the training process while striking a balance in the exploration of RL. The significance of the variance value in RL training is crucial for incorporating physical commonsense. Therefore, we conducted ablation experiments on sigma, ranging from 0.1 to 0.3, as shown in Table \ref{table:sigma}. It can be observed that ReinDiffuse achieves the best FID performance when $\sigma$ is set to 0.15. Consequently, we adopt $\sigma$ = 0.15 as the default value for our RL training. Interestingly, the physical performance does not exhibit a consistent increase or decrease with the variation of sigma. In fact, when we have very large sigma values (approaching 1), the model fails to converge. This indicates that sigma plays a crucial role in the updates for RL.

\section{Conclusion}
\label{sec:conclusion}

In this paper, we propose ReinDiffuse to enable direct reinforcement learning (RL) training of Motion Diffusion Model (MDM) without modifying the network structure. To address the issue of generating physically implausible motion, we employ frame-wise physical rewards for RL fine-tuning of MDM. The results demonstrate outstanding capabilities in generating physically plausible motion. Extensive qualitative and quantitative experiments validate the effectiveness and robustness of our method. 

{\bf Limitations:} Our limitations lie in the requirement to formulate a suitable physical plausibility reward function for each physical problem, which may result in a considerable workload for reward design when confronted with a multitude of physical problems. Additionally, our physical rewards are calculated based on joint locations, and some fine-grained physical issues may only arise when rendered as a mesh formats (e.g., interpenetration between hand and body). Rendering mesh data during training is extremely time-consuming. Due to computational resource constraints, we are unable to calculate physical plausibility rewards based on mesh data. 

{\bf Future Works:} We have only explored using physical rewards to enhance the model's physical performance. Training with semantically related rewards holds great promise.

{\small
\bibliographystyle{ieee_fullname}
\bibliography{egbib}

\begin{thebibliography}{10}\itemsep=-1pt

\bibitem{alexanderson2022listen}
Simon Alexanderson, Rajmund Nagy, Jonas Beskow, and Gustav~Eje Henter.
\newblock Listen, denoise, action! audio-driven motion synthesis with diffusion models.
\newblock {\em arXiv preprint arXiv:2211.09707}, 2022.

\bibitem{TEACH:3DV:2022}
Nikos Athanasiou, Mathis Petrovich, Michael~J. Black, and G\"{u}l Varol.
\newblock Teach: Temporal action compositions for 3d humans.
\newblock In {\em International Conference on 3D Vision (3DV)}, September 2022.

\bibitem{black2023ddpo}
Kevin Black, Michael Janner, Yilun Du, Ilya Kostrikov, and Sergey Levine.
\newblock Training diffusion models with reinforcement learning.
\newblock {\em arXiv preprint arXiv:2305.13301}, 2023.

\bibitem{chao2021learning}
Yu-Wei Chao, Jimei Yang, Weifeng Chen, and Jia Deng.
\newblock Learning to sit: Synthesizing human-chair interactions via hierarchical control.
\newblock In {\em Proceedings of the AAAI Conference on Artificial Intelligence}, volume~35, pages 5887--5895, 2021.

\bibitem{chen2022mld}
Xin Chen, Biao Jiang, Wen Liu, Zilong Huang, Bin Fu, Tao Chen, and Gang Yu.
\newblock Executing your commands via motion diffusion in latent space.
\newblock In {\em Proceedings of the IEEE/CVF Conference on Computer Vision and Pattern Recognition (CVPR)}, pages 18000--18010, June 2023.

\bibitem{chen2022learning}
Xin Chen, Zhuo Su, Lingbo Yang, Pei Cheng, Lan Xu, Bin Fu, and Gang Yu.
\newblock Learning variational motion prior for video-based motion capture.
\newblock {\em arXiv preprint arXiv:2210.15134}, 2022.

\bibitem{fan2023dpok}
Ying Fan, Olivia Watkins, Yuqing Du, Hao Liu, Moonkyung Ryu, Craig Boutilier, Pieter Abbeel, Mohammad Ghavamzadeh, Kangwook Lee, and Kimin Lee.
\newblock Dpok: Reinforcement learning for fine-tuning text-to-image diffusion models.
\newblock {\em arXiv preprint arXiv:2305.16381}, 2023.

\bibitem{Guo_2022_CVPR_humanml3d}
Chuan Guo, Shihao Zou, Xinxin Zuo, Sen Wang, Wei Ji, Xingyu Li, and Li Cheng.
\newblock Generating diverse and natural 3d human motions from text.
\newblock In {\em Proceedings of the IEEE/CVF Conference on Computer Vision and Pattern Recognition (CVPR)}, pages 5152--5161, June 2022.

\bibitem{Guo_2022_CVPR_t2m}
Chuan Guo, Shihao Zou, Xinxin Zuo, Sen Wang, Wei Ji, Xingyu Li, and Li Cheng.
\newblock Generating diverse and natural 3d human motions from text.
\newblock In {\em Proceedings of the IEEE/CVF Conference on Computer Vision and Pattern Recognition (CVPR)}, pages 5152--5161, June 2022.

\bibitem{guo2022t2m}
Chuan Guo, Shihao Zou, Xinxin Zuo, Sen Wang, Wei Ji, Xingyu Li, and Li Cheng.
\newblock Generating diverse and natural 3d human motions from text.
\newblock In {\em Proceedings of the IEEE/CVF Conference on Computer Vision and Pattern Recognition}, pages 5152--5161, 2022.

\bibitem{guo2022tm2t}
Chuan Guo, Xinxin Zuo, Sen Wang, and Li Cheng.
\newblock Tm2t: Stochastic and tokenized modeling for the reciprocal generation of 3d human motions and texts.
\newblock In {\em Computer Vision--ECCV 2022: 17th European Conference, Tel Aviv, Israel, October 23--27, 2022, Proceedings, Part XXXV}, pages 580--597. Springer, 2022.

\bibitem{han2023hutumotion}
Gaoge Han, Shaoli Huang, Mingming Gong, and Jinglei Tang.
\newblock Hutumotion: Human-tuned navigation of latent motion diffusion models with minimal feedback.
\newblock {\em arXiv preprint arXiv:2312.12227}, 2023.

\bibitem{hassan2023synthesizing}
Mohamed Hassan, Yunrong Guo, Tingwu Wang, Michael Black, Sanja Fidler, and Xue~Bin Peng.
\newblock Synthesizing physical character-scene interactions.
\newblock In {\em ACM SIGGRAPH 2023 Conference Proceedings}, pages 1--9, 2023.

\bibitem{heusel2017gansfid}
Martin Heusel, Hubert Ramsauer, Thomas Unterthiner, Bernhard Nessler, and Sepp Hochreiter.
\newblock Gans trained by a two time-scale update rule converge to a local nash equilibrium.
\newblock {\em Advances in neural information processing systems}, 30, 2017.

\bibitem{jiang2023motiongpt}
Biao Jiang, Xin Chen, Wen Liu, Jingyi Yu, Gang Yu, and Tao Chen.
\newblock Motiongpt: Human motion as a foreign language.
\newblock {\em arXiv preprint arXiv:2306.14795}, 2023.

\bibitem{kakade2002approximately}
Sham Kakade and John Langford.
\newblock Approximately optimal approximate reinforcement learning.
\newblock In {\em Proceedings of the Nineteenth International Conference on Machine Learning}, pages 267--274, 2002.

\bibitem{karunratanakul2023GMD}
Korrawe Karunratanakul, Konpat Preechakul, Supasorn Suwajanakorn, and Siyu Tang.
\newblock Guided motion diffusion for controllable human motion synthesis.
\newblock In {\em Proceedings of the IEEE/CVF International Conference on Computer Vision}, pages 2151--2162, 2023.

\bibitem{katehakis1987multi}
Michael~N Katehakis and Arthur~F Veinott~Jr.
\newblock The multi-armed bandit problem: decomposition and computation.
\newblock {\em Mathematics of Operations Research}, 12(2):262--268, 1987.

\bibitem{kingma2013autovae}
Diederik~P Kingma and Max Welling.
\newblock Auto-encoding variational bayes.
\newblock {\em arXiv preprint arXiv:1312.6114}, 2013.

\bibitem{lee2019dancing}
Hsin-Ying Lee, Xiaodong Yang, Ming-Yu Liu, Ting-Chun Wang, Yu-Ding Lu, Ming-Hsuan Yang, and Jan Kautz.
\newblock Dancing to music.
\newblock {\em Advances in neural information processing systems}, 32, 2019.

\bibitem{li2022danceformer}
Buyu Li, Yongchi Zhao, Shi Zhelun, and Lu Sheng.
\newblock Danceformer: Music conditioned 3d dance generation with parametric motion transformer.
\newblock In {\em Proceedings of the AAAI Conference on Artificial Intelligence}, volume~36, pages 1272--1279, 2022.

\bibitem{li2023object}
Jiaman Li, Jiajun Wu, and C~Karen Liu.
\newblock Object motion guided human motion synthesis.
\newblock {\em arXiv preprint arXiv:2309.16237}, 2023.

\bibitem{li2021ai}
Ruilong Li, Shan Yang, David~A Ross, and Angjoo Kanazawa.
\newblock Ai choreographer: Music conditioned 3d dance generation with aist++.
\newblock In {\em Proceedings of the IEEE/CVF International Conference on Computer Vision}, pages 13401--13412, 2021.

\bibitem{lu2010contextual}
Tyler Lu, D{\'a}vid P{\'a}l, and Martin P{\'a}l.
\newblock Contextual multi-armed bandits.
\newblock In {\em Proceedings of the Thirteenth international conference on Artificial Intelligence and Statistics}, pages 485--492. JMLR Workshop and Conference Proceedings, 2010.

\bibitem{makoviychuk2021isaac}
Viktor Makoviychuk, Lukasz Wawrzyniak, Yunrong Guo, Michelle Lu, Kier Storey, Miles Macklin, David Hoeller, Nikita Rudin, Arthur Allshire, Ankur Handa, et~al.
\newblock Isaac gym: High performance gpu-based physics simulation for robot learning.
\newblock {\em arXiv preprint arXiv:2108.10470}, 2021.

\bibitem{ouyang2022traininginstructgpt}
Long Ouyang, Jeffrey Wu, Xu Jiang, Diogo Almeida, Carroll Wainwright, Pamela Mishkin, Chong Zhang, Sandhini Agarwal, Katarina Slama, Alex Ray, et~al.
\newblock Training language models to follow instructions with human feedback.
\newblock {\em Advances in Neural Information Processing Systems}, 35:27730--27744, 2022.

\bibitem{pan2023synthesizing}
Liang Pan, Jingbo Wang, Buzhen Huang, Junyu Zhang, Haofan Wang, Xu Tang, and Yangang Wang.
\newblock Synthesizing physically plausible human motions in 3d scenes.
\newblock {\em arXiv preprint arXiv:2308.09036}, 2023.

\bibitem{pavlakos2019expressive}
Georgios Pavlakos, Vasileios Choutas, Nima Ghorbani, Timo Bolkart, Ahmed~AA Osman, Dimitrios Tzionas, and Michael~J Black.
\newblock Expressive body capture: 3d hands, face, and body from a single image.
\newblock In {\em Proceedings of the IEEE/CVF conference on computer vision and pattern recognition}, pages 10975--10985, 2019.

\bibitem{petrovich22temos}
Mathis Petrovich, Michael~J. Black, and G{\"u}l Varol.
\newblock {TEMOS}: Generating diverse human motions from textual descriptions.
\newblock In {\em European Conference on Computer Vision ({ECCV})}, 2022.

\bibitem{pinto2023tuning}
Andr{\'e}~Susano Pinto, Alexander Kolesnikov, Yuge Shi, Lucas Beyer, and Xiaohua Zhai.
\newblock Tuning computer vision models with task rewards.
\newblock {\em arXiv preprint arXiv:2302.08242}, 2023.

\bibitem{Plappert2016kit}
Matthias Plappert, Christian Mandery, and Tamim Asfour.
\newblock The kit motion-language dataset.
\newblock {\em Big Data}, 4(4):236--252, dec 2016.

\bibitem{rempe2021humor}
Davis Rempe, Tolga Birdal, Aaron Hertzmann, Jimei Yang, Srinath Sridhar, and Leonidas~J Guibas.
\newblock Humor: 3d human motion model for robust pose estimation.
\newblock In {\em Proceedings of the IEEE/CVF international conference on computer vision}, pages 11488--11499, 2021.

\bibitem{schulman2017proximal}
John Schulman, Filip Wolski, Prafulla Dhariwal, Alec Radford, and Oleg Klimov.
\newblock Proximal policy optimization algorithms.
\newblock {\em arXiv preprint arXiv:1707.06347}, 2017.

\bibitem{mdm2022human}
Guy Tevet, Sigal Raab, Brian Gordon, Yonatan Shafir, Amit~H Bermano, and Daniel Cohen-Or.
\newblock Human motion diffusion model.
\newblock {\em arXiv preprint arXiv:2209.14916}, 2022.

\bibitem{tseng2022edge}
Jonathan Tseng, Rodrigo Castellon, and C~Karen Liu.
\newblock Edge: Editable dance generation from music.
\newblock {\em arXiv preprint arXiv:2211.10658}, 2022.

\bibitem{van2017neuralvqvae}
Aaron Van Den~Oord, Oriol Vinyals, et~al.
\newblock Neural discrete representation learning.
\newblock {\em Advances in neural information processing systems}, 30, 2017.

\bibitem{wang2021synthesizing}
Jiashun Wang, Huazhe Xu, Jingwei Xu, Sifei Liu, and Xiaolong Wang.
\newblock Synthesizing long-term 3d human motion and interaction in 3d scenes.
\newblock In {\em Proceedings of the IEEE/CVF Conference on Computer Vision and Pattern Recognition}, pages 9401--9411, 2021.

\bibitem{williams1992simpleREINFORCE}
Ronald~J Williams.
\newblock Simple statistical gradient-following algorithms for connectionist reinforcement learning.
\newblock {\em Machine learning}, 8:229--256, 1992.

\bibitem{wu2023fine}
Zeqiu Wu, Yushi Hu, Weijia Shi, Nouha Dziri, Alane Suhr, Prithviraj Ammanabrolu, Noah~A Smith, Mari Ostendorf, and Hannaneh Hajishirzi.
\newblock Fine-grained human feedback gives better rewards for language model training.
\newblock {\em arXiv preprint arXiv:2306.01693}, 2023.

\bibitem{xie2023omnicontrol}
Yiming Xie, Varun Jampani, Lei Zhong, Deqing Sun, and Huaizu Jiang.
\newblock Omnicontrol: Control any joint at any time for human motion generation.
\newblock {\em arXiv preprint arXiv:2310.08580}, 2023.

\bibitem{xu2023imagereward}
Jiazheng Xu, Xiao Liu, Yuchen Wu, Yuxuan Tong, Qinkai Li, Ming Ding, Jie Tang, and Yuxiao Dong.
\newblock Imagereward: Learning and evaluating human preferences for text-to-image generation.
\newblock {\em arXiv preprint arXiv:2304.05977}, 2023.

\bibitem{yan2019convolutional}
Sijie Yan, Zhizhong Li, Yuanjun Xiong, Huahan Yan, and Dahua Lin.
\newblock Convolutional sequence generation for skeleton-based action synthesis.
\newblock In {\em Proceedings of the IEEE/CVF International Conference on Computer Vision}, pages 4394--4402, 2019.

\bibitem{yuan2023physdiff}
Ye Yuan, Jiaming Song, Umar Iqbal, Arash Vahdat, and Jan Kautz.
\newblock Physdiff: Physics-guided human motion diffusion model.
\newblock In {\em Proceedings of the IEEE/CVF International Conference on Computer Vision}, pages 16010--16021, 2023.

\bibitem{zhang2023diffmotion}
Fan Zhang, Naye Ji, Fuxing Gao, and Yongping Li.
\newblock Diffmotion: Speech-driven gesture synthesis using denoising diffusion model.
\newblock In {\em International Conference on Multimedia Modeling}, pages 231--242. Springer, 2023.

\bibitem{zhang2023t2mgpt}
Jianrong Zhang, Yangsong Zhang, Xiaodong Cun, Shaoli Huang, Yong Zhang, Hongwei Zhao, Hongtao Lu, and Xi Shen.
\newblock T2m-gpt: Generating human motion from textual descriptions with discrete representations.
\newblock {\em arXiv preprint arXiv:2301.06052}, 2023.

\bibitem{zhang2022motiondiffuse}
Mingyuan Zhang, Zhongang Cai, Liang Pan, Fangzhou Hong, Xinying Guo, Lei Yang, and Ziwei Liu.
\newblock Motiondiffuse: Text-driven human motion generation with diffusion model.
\newblock {\em arXiv preprint arXiv:2208.15001}, 2022.

\bibitem{zhang2023remodiffuse}
Mingyuan Zhang, Xinying Guo, Liang Pan, Zhongang Cai, Fangzhou Hong, Huirong Li, Lei Yang, and Ziwei Liu.
\newblock Remodiffuse: Retrieval-augmented motion diffusion model.
\newblock {\em arXiv preprint arXiv:2304.01116}, 2023.

\bibitem{zhang2020perpetual}
Yan Zhang, Michael~J Black, and Siyu Tang.
\newblock Perpetual motion: Generating unbounded human motion.
\newblock {\em arXiv preprint arXiv:2007.13886}, 2020.

\bibitem{zhao2020bayesian}
Rui Zhao, Hui Su, and Qiang Ji.
\newblock Bayesian adversarial human motion synthesis.
\newblock In {\em Proceedings of the IEEE/CVF Conference on Computer Vision and Pattern Recognition}, pages 6225--6234, 2020.

\end{thebibliography}
}

\end{document}